\documentclass[conference]{IEEEtran}
\IEEEoverridecommandlockouts
\usepackage{cite}
\usepackage{amsmath,amssymb,amsfonts}
\usepackage{algorithmic}
\usepackage{graphicx}
\usepackage{textcomp}
\usepackage{xcolor}
\def\BibTeX{{\rm B\kern-.05em{\sc i\kern-.025em b}\kern-.08em
    T\kern-.1667em\lower.7ex\hbox{E}\kern-.125emX}}
\begin{document}
\IEEEpubid{\makebox[\columnwidth]{\copyright~2019 IEEE \hfill} \hspace{\columnsep}\makebox[\columnwidth]{ }}

\title{3D Anchor-Free Lesion Detector on Computed Tomography Scans\\
\thanks{* indicates equal contribution.}
}

\author{\IEEEauthorblockN{Ning Zhang*, Dechun Wang*, Xinzi Sun, Pengfei Zhang, Chenxi Zhang, Yu Cao, Benyuan Liu}
\IEEEauthorblockA{\textit{Department Of Computer Science} \\
\textit{University of Massachusetts Lowell}\\
Lowell, MA \\
\{ning\_zhang, dechun\_wang, xinzi\_sun, pengfei\_zhang,  chenxi\_zhang\}@student.uml.edu\\ \{ycao, bliu\}@cs.uml.edu}}

\maketitle
\begin{abstract}
Lesions are injuries and abnormal tissues in the human body. Detecting lesions in 3D Computed Tomography (CT) scans can be time-consuming even for very experienced physicians and radiologists. In recent years, CNN based lesion detectors have demonstrated huge potentials. Most of current state-of-the-art lesion detectors employ anchors to enumerate all possible bounding boxes with respect to the dataset in process. This anchor mechanism greatly improves the detection performance while also constraining the generalization ability of detectors. In this paper, we propose an anchor-free lesion detector. The anchor mechanism is removed and lesions are formalized as single keypoints. By doing so, we witness a considerable performance gain in terms of both accuracy and inference speed compared with the anchor-based baseline.
\end{abstract}


\section{Introduction}
Computed Tomography (CT) scans capture inner details of the human body by emitting a series of narrow beam of X-ray. The X-ray absorption differs much across different tissues of the human body. This provides a way for physicians and radiologists to examine across different healthy organs as well as abnormal lesions. 

Lesions are injuries and abnormal tissues. They can locate in different organs such as lungs, livers, abdomens, bones, etc. and are often the early stage manifestations of fatal diseases such as cancers and tuberculosis. Detecting lesions at their early stages are believed to improve the cure rate and survival rate. Compared with healthy tissues, lesions often present distinctive visual properties in CT scans. For instance, pulmonary (lung) nodules (also referred to as coin lesions) are often small round or oval-shaped with an isolated absorption of X-ray (measured by Hounsfield Unit). With these properties, it is possible for machines to detect lesions automatically from CT scans. 

Before the advent of CNN, people mainly resorted to different types of morphology features such as Shape Index (SI), Curvedness (CV) \cite{henschke2002ct,murphy2009large,jacobs2014automatic} and other well designed features \cite{setio2017validation} for this this task. These features are devised fully based on human knowledge and have long been playing an important role. However, the limit is obvious as it is not easy for a human to enumerate all lesion appearance in the real world. One direct result is the relatively low recall rate.

In recent years, this feature engineering process is replaced by deep convolutional neural networks \cite{he2016deep, huang2017densely, Zhang:2017:IMR:3123266.3123332,7545842,7545830,alcantara2017improving,7255222,sun2019People} where rich features are learned automatically. Most of CNN based lesion detectors \cite{yan20183d, zhongliuxie2018, liao2017evaluate} adopt the anchor mechanism to enumerate all possible bounding box templates (anchors) with respect to the dataset in terms of aspect ratio and size. These anchors can greatly improve recall but also cause massive false positives. Moreover, these massive false positives can exert huge pressure on Non-Maximum Suppression, making inference slow. Another issue with the anchor mechanism is that the anchor configuration must fit well to the characteristics of the dataset. Otherwise, a big degradation in performance can happen. This issue becomes more severe when objects under concern are very small \cite{eggert2017closer}.

The anchor-free idea seems to fit well with our task. One reason for this is to save the effort in finding the best anchor settings when it is adopted to different datasets. The other reason lies in the observation that lesions in 3D CT scans do not overlap with each other. Thus, we think overlapped anchors may not be necessary for our task.

Our major contribution in this paper lies in that we the first to propose a 3D anchor-free architecture for the lesion detection task. 


\begin{figure*}[t]
	\begin{center}
		\includegraphics[width=0.9\linewidth,height=1.8in]{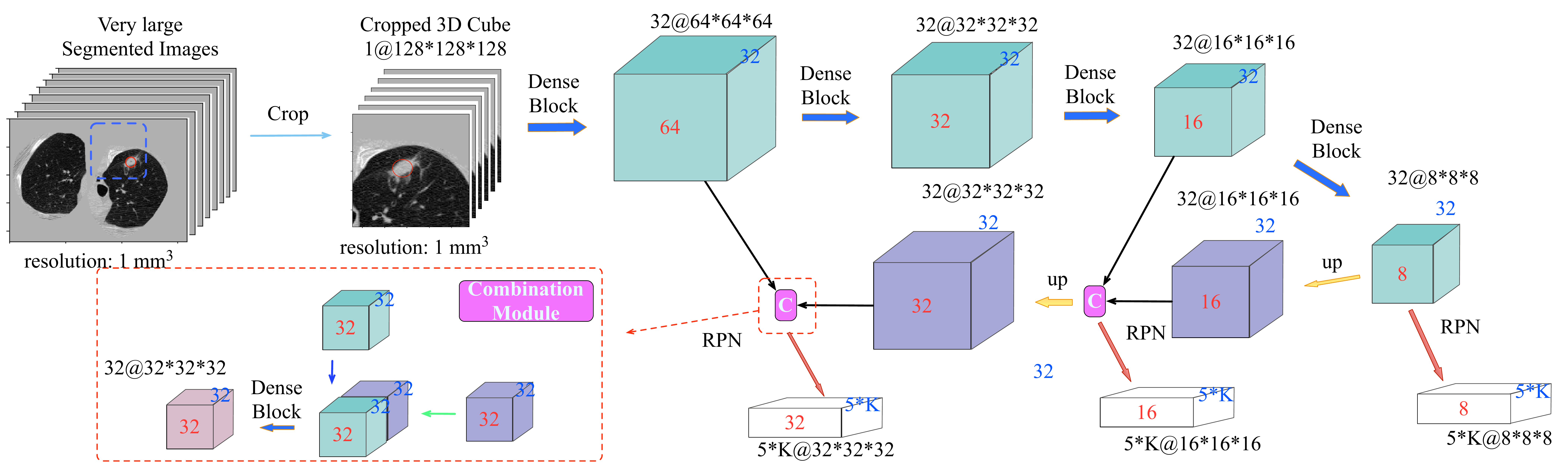}
	\end{center}
	\caption{The architecture of the whole network. The network is of a ``U" structure and consists of an upstream and a downstream pathway. Upstream and downstream features would be concatenated before being forwarded to the next layer. Detection heads are attached to the combined features. Note that $K=1$ for the anchor-free setting while $K=3$ for the anchor-based setting.} 
	\label{fig:network}
\end{figure*}
\section{Related Work}
\subsection{Lesion Detection}
Pulmonary Nodule detection has been well studied for years. Liao et. al. \cite{liao2017evaluate} proposed a 3D U-net \cite{ronneberger2015u} for the nodule detection task. This U-net contains upstream and downstream pathways which are similar to Feature Pyramid Network \cite{lin2017feature}. However, the major difference lies in the upsampling operations in the top-down pathway and how upstream features and downstream features are combined. In particular, U-net adopts the transposed convolutions while FPN emploies parameter-free interpolations. To combines upstream and downstream features, the former leverages concatenation while the latter uses element-wise addition. Note that CT scans are usually too large. Thus, only sub-cubes of the original scans (typically 128$\times$128$\times$128, $\sim$1/6 of the whole image) are fed in as the input. Zhu et. al \cite{zhu2018deeplung} replaced the ResNet Building block with the DualNet block and a better performance is reported. 

It is reported in \cite{yan20183d} that 3D CNN may perform poorly when only part of the CT scan is provided such as the DeepLesion dataset \cite{Yan_2018_CVPR}. Therefore, instead of employing 3D CNNs, Yan et. al \cite{yan20183d} proposed a 3D context enhanced 2D CNN for the general lesion detection task. The 3D context is achieved by stacking CNN features extracted from neighboring slices. One drawback is that this approach requires the ``key slice" known in advance while in real application scenarios the ``key slice" is agnostic, which limits its practicability.

\subsection{Anchor-Free Detectors}
Recently, anchor-free Detectors have demonstrated great potentials and gained much attention \cite{zhou2019keypoint, DBLP:cornernet}. The main motivation is to remove the hassle of devising the best anchors (bounding box templates) for the dataset in process. In these anchor-free detectors, objects are represented as either a pair of keypoints at the corners (top-left corners and bottom-right corners) \cite{DBLP:cornernet} or single keypoints in the center \cite{zhou2019keypoint}. To better detect these keypoints, enriching context information is proved to be critical. To this end, a special corner pooling and a center pooling are adopted. However, one issue with these pooling operations is that they are very slow.

In our paper, we do not employ these powerful yet slow pooling operations. One reason comes from the object size. In our case, lesions are often very small and the receptive field of the detection head is relatively large compared to the object size. Moreover, lesions do not overlap with each other, making this pooling less necessary.

\section{Our Approach}

\subsection{Network Architecture}
We employ a U-net structure built upon DenseNet \cite{huang2017densely} building blocks. Our detector is singe-stage and each feature map is attached with a detection head. These heads are of an identical structure while they are independent with each other (no sharing). The whole network architecture is illustrated in Fig. \ref{fig:network}. In this paper, proposals in anchor-based and anchor-free settings are both encoded as 5-element vectors $\{score, coord_x, coord_y, coord_z, diameter\}$. This encoding method is also adopted in field practice by physicians and radiologists when locating lesions. Therefore, the output channels of the detection heads (as shown in Fig. \ref{fig:network}) are $5\times K$ where $K=3$ and $K=1$ for the anchor-based and anchor-free setting respectively. 

\subsection{Ground Truth Assignment}
In our anchor-free design, we formulate a object as a center keypoint. We define a positive cube and non-negative cube for each grouth truth box. Center points locating inside of the positive cube, outside of the non-negative cube and in between will be assigned as positive, negative and ignored respectively. In particular, consider an object $b = (c_x , c_y, c_z, d)$, where $(c_x, c_y, c_z)$ is the centroid; $d$ is the diameter. Suppose this object is assigned to feature map $F_l$ with size $m$ and stride $s_l$, we have $m^3$ (assuming the input is a cube for simplicity) center points ${P}=\{x_i,y_j,z_k\}^{m}_{i,j,k=1}$ where $x_i = s_l \times i$, $y_j = s_l \times j$, $z_k = s_l \times k$. The positive cube $b^l_p = (c_x,c_y,c_z, d^l_p)$ and non-negative cube $b^l_n = (c_x,c_y,c_z, d^l_n)$, where $d^l_p =  \varepsilon_p d, ,d^l_n =  \varepsilon_n d$. Center points will be marked as positive if $p_{ijk} \in b^l_p $, negative if $p_{ijk} \notin b^l_n$ and ignored if $p_{ijk} \notin b^l_p \land p_{ijk} \in b^l_n $. We use $ \varepsilon_p=0.8$ and $\varepsilon_n=1.2 $ in this paper. Note that the anchor-based baseline adopts the standard IoU based algorithm to label individual anchors. 


\subsection{Training Loss}
The training loss can be divided into the classification part and the localization part: $ \mathcal{L} = \mathcal{L}_{cls}+ \mathcal{L}_{loc}$. For the classification part, we use Focal Loss \cite{lin2017focal} for negative samples and Cross Entropy for positive samples. In addition, we follow \cite{DBLP:cornernet} and penalize positive center points with a unnormalized Gaussian determined by its Euclidean distance to the ground truth centroid and the size of the object. Formally, given an object $g = (c^g_x, c^g_y, c^g_z, d^g)$ and a positive center point $ p_{ijk} = (c^i_x, c^j_y, c^k_z)$, the weight $ \psi_{p_{ijk}} $ is defined as:
\begin{equation}
\psi_{p_{ijk}}=\exp\big( -\dfrac{(c^g_x-c^i_x)^2 +( c^g_y-c^j_y)^2+( c^g_z-c^k_z)^2}{2\alpha(d^g)^2}\big).
\end{equation}
We use $ \alpha = 1 $ in this paper. After this, we have the following loss defined for the classification end:
\begin{equation}
    \mathcal{L}_F(p_t) = -\alpha_t(1-p_t)^\gamma log(p_t)
\label{vfocal}
\end{equation}
\begin{equation}
\begin{aligned}
    \mathcal{L}_{cls} = 1/N_{pos}\sum_j^{N_{neg}}\mathcal{L}_F(p_{t_j}) 
    + 1/N_{pos}\sum_i^{N_{pos}}\psi_{i}\mathcal{L}_{CE}(p_{i}),
\end{aligned}
\label{tfocal}
\end{equation}
where $p_t = (p)^y(1-p)^{(1-y)}$, $y \in \{0,1\}$ is the ground truth. 

For the localization part, the offset targets are encoded with the stride of feature maps instead of the anchors. More formally, the offset target $ v^l_{g,p_{ijk}} =[\Delta c^{l}_x,\Delta c^{l}_y, \Delta c^{l}_z, \Delta d^l]$ on feature map $F_l$ with stride $s_l$ is defined as follows:
\begin{equation}
\begin{aligned}
\Delta c^{l}_x &= \dfrac{(c^g_x-c^p_x)}{s_l}, & \Delta c^{l}_y &= \dfrac{(c^g_y-c^p_y)}{s_l}\\
\Delta c^{l}_z &= \dfrac{(c^b_z-c^p_z)}{s_l}, & \Delta d^l &=\log(\dfrac{d^g}{s_l}).
\end{aligned}
\end{equation}
For the localization part we adopt the Smooth L1 loss.

\begin{table*}
\caption{Sensitivity (\%), FROC score and Inference Time (s/scan) on the DeepLesion dataset. 
Note that one may not directly compare performance with \cite{yan20183d} because of the different task settings (2D vs 3D).}
    \centering
    \begin{tabular}{c|c|c|c|c|c|c|c|c|c} 
    FPs per image & 0.5 & 1 & 2 & 4 & 8 & 16 & Avg. & FROC & Inference Time \\ \hline \hline
    3DCE, 27 slices \cite{yan20183d} & 62.48 &73.37 &80.70 &85.65 &89.09 &91.06 & 80.39 & - & -\\ \hline
    Anchor-Based RPN  & 65.74 &73.89 &80.99 &86.56 & 91.40 & 94.40 & 82.17 & 0.708 & 1.95s \\
    Anchor-Free RPN   & \textbf{68.73} &\textbf{77.10}  &\textbf{83.54}  &\textbf{88.12} 
    &\textbf{91.94}  &\textbf{94.62} & \textbf{84.01} & \textbf{0.735} & \textbf{1.74s} \\\hline
 \end{tabular}
\label{tb:deeplesion}   
\end{table*}

\begin{table*}
\caption{Sensitivity@4 (\%) w.r.t lesion type and size.  
Types include lung (LU), mediastinum (ME), liver (LV), soft tissue (ST), pelvis (PV), abdomen (AB), kidney (KD), and bone (BN), respectively. ``$<$10", ``10-30" and ``$>$30" represent lesion diameter ranges (mm).}
    \centering
    \begin{tabular}{c|cccccccc|ccc} 
    Model &LU &ME &LV &ST &PV &AB &KD &BN & $<$10 &10-30 & $>$30  \\ \hline \hline
    3DCE, 27 slices \cite{yan20183d} & 89 &88 &90 &74 &84 &84 &82 &75 &80 &87 &84 \\ \hline
    Anchor-Based RPN & 91 &\textbf{88} &87 &80 &85 &80 &80 &\textbf{69} &82 &\textbf{88} &80   \\
    Anchor-Free RPN & \textbf{93} &88 &\textbf{91} &\textbf{85} &\textbf{86} &\textbf{83} & \textbf{80} &65 &\textbf{83} & 87 &\textbf{88} \\\hline
 \end{tabular}
\label{tb:deeplesion_subtype}
\end{table*}

\begin{figure}[t]
	\begin{center}
		\includegraphics[width=1.0\linewidth,height=1.3in]{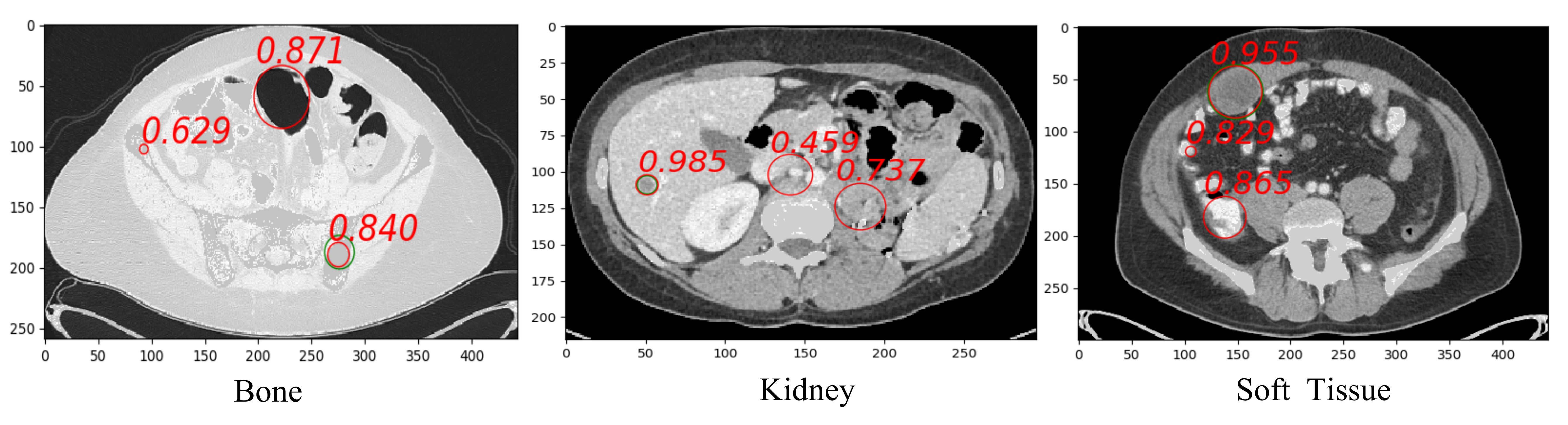}
	\end{center}
	\caption{Visualization of lesion types: bone, kidney and soft tissue. Red and green circles are predicted and ground truth boxes respectively. It may not be clear that the box 0.985 in kidney and 0.955 in soft tissue nearly fit perfectly with the ground truth. } 
	\label{fig:sub_type}
\end{figure}

\section{Experiments}
We conduct experiments on the DeepLesion \cite{Yan_2018_CVPR} dataset. This dataset is designed for general lesion detection with various types of lesion, including lung, mediastinum, liver, soft tissue, pelvis, abdomen, kidney, and bone. This dataset contains 10,594 CT studies from 4,427 unique patients with 32,735 annotated lesions. The official training, validation and testing set containing 22,901, 4,887, 4,912 lesions respectively (noisy annotations are removed). Note that, for each lesion, DeepLesion only provides a 60mm Z-context chunk centered with the annotated slice (key slice).

Primary attempts in \cite{yan20183d} indicated that 3D CNN may not work well with the DeepLesion dataset. We think the reason is 3-fold: (1) out-bounding large lesions ($\geq$ 48mm, $\sim$11\% of all lesions) make it hard for localization (both the center position and size). (2)  z-coordinate may not be accurate. As lesions are only annotated on a single center slice, when the slice interval is large the annotations would be inaccurate. (3) small lesions tend to be assigned with over-sized bounding boxes. This again introduces noises for the 3D CNN to regress the size.


\subsection{Data Pre-processing}
We rescale (by interpolation) the 3D CT scans to an isotropic resolution (1mm in all directions). The large mass of black borders of the image is removed by simple value clipping. During training, random crops of a size 64$\times$128$\times$128 (padding 0 when necessary) are fed into the network during training while in testing, a sliding window style cropping strategy is adopted. Detection results on these sub-crop pieces will be assembled to form the integral result. The 2D annotations are approximately converted to 3D ones with the form of \{X, Y, Z, Diameter\}.  


\subsection{Training and Testing}
Unlike \cite{yan20183d}, our tasks remain in the 3D object detection regime. We train the 3D CNN both w/ and w/o anchors. In anchor-based model, we configure 3 anchors for each feature scale (stride 4, 8, 16) which are \{3.0, 5.0, 7.0\}, \{10.0, 13.0, 17.0\},\{22.0, 30.0, 40.0\} respectively. During training, very large lesions ($\geq$ 48mm, $\sim$11\% of the training data) are removed because of the regression issues. Primary attempts show that if these large lesions are included, the training process would suffer from a slow convergence and oscillating losses. During testing, very large lesions are included.

\subsection{Evaluation}
We detect lesion on each 60-mm z-axis CT image chunk. We use the free receiver operating characteristic (FROC) score to evaluate the performance following the same protocol of LUNA16 challenge \cite{setio2017validation}. This FROC score is approximated by the average recall at 7 false positive rates: 1/8, 1/4, 1/2, 1, 2, 4, and 8 False Positive per scan. In our case, one predicted box would be counted as a True Positive if its centroid is located in the mass of ground truth. In other words, the distance between the proposed and the real centroid is less than the radius of ground truth. 

\subsection{Overall Performance}
As we can see from Table \ref{tb:deeplesion}, 3D CNNs work well with this task. Note that in \cite{yan20183d} evaluate their model at the key slice while we are detecting lesions without knowing the key slice in advance. Therefore, we argue our task settings are more challenging. In addition, our anchor-free RPN outperforms the anchor-based RPN in terms of both accuracy and inference speed. Our inference time is evaluated with one Nvidia Telsa K80. 

\subsection{Performance w.r.t. Lesion Type and Size}
Following \cite{yan20183d}, we also report the performance with respect to lesion type and diameter. All results are summarized in Table \ref{tb:deeplesion_subtype}. We can find that our 3D models (w/ and w/o anchors) do not perform well for bone and kidney lesions. On the other hand, our approach experiences no significant performance drop as \cite{yan20183d} when detecting ``Soft Tissue" lesions. We visualize these types in Fig. \ref{fig:sub_type}. Another observation is that the anchor-free design seems to be more tolerable to very large lesions than the anchor-based counterpart (``$>30$" in Table \ref{tb:deeplesion_subtype} ). Again, we stress the point that one may not directly compare our results with \cite{yan20183d}.

\section{Conclusions}
Our anchor-free design works well with the general lesion detection task in terms of both accuracy and inference speed. Compared with the anchor-based design, the anchor-free design is more robust to large lesions (potentially reaching boundaries). Even though we cannot directly compared with \cite{yan20183d}. We argue that our model can work with the key slice agnostic scenarios, which is more practical for real applications.





\bibliographystyle{IEEEtran}
\bibliography{references}

\end{document}